\documentclass[10pt,twocolumn,letterpaper]{article}

\usepackage{cvpr}              
\usepackage{makecell}
\usepackage{diagbox}
\usepackage{graphicx}
\usepackage{marvosym}
\usepackage{url}            
\usepackage{booktabs}       
\usepackage{amsfonts}       
\usepackage{nicefrac}       
\usepackage{microtype}      
\usepackage{xcolor}         
\usepackage{wrapfig} 

\usepackage{adjustbox}
\usepackage{diagbox}
\usepackage{multirow}
\usepackage{makecell}

\usepackage{times}
\usepackage{epsfig}
\usepackage{graphicx}
\usepackage{amsmath}
\usepackage{amssymb}
\usepackage{colortbl}
\usepackage{xspace}

\usepackage{ragged2e}
\usepackage{float}
\usepackage{amssymb}  
\usepackage[accsupp]{axessibility}

\usepackage{gensymb}
\usepackage{fontawesome5}
\usepackage{url}            
\usepackage{booktabs}       
\usepackage{amsfonts}       
\usepackage{nicefrac}       
\usepackage{microtype}      
\usepackage{xcolor}         
\usepackage{wrapfig} 

\usepackage{adjustbox}
\usepackage{diagbox}
\usepackage{multirow}
\usepackage{makecell}
\usepackage{times}
\usepackage{epsfig}
\usepackage{graphicx}
\usepackage{amsmath}
\usepackage{amssymb}
\usepackage{colortbl}
\usepackage{xspace}

\usepackage{ragged2e}
\usepackage{float}
\usepackage{amssymb}  
\usepackage[accsupp]{axessibility}
\usepackage{cuted}
\usepackage{hyperref}
\definecolor{cvprblue}{rgb}{0.21,0.49,0.74}
\def\paperID{34758} 
\def\confName{CVPR}
\def\confYear{2026}

\title{AT-VLA: Adaptive Tactile Injection for Enhanced Feedback Reaction in Vision-Language-Action Models}

\author{
Xiaoqi Li \textsuperscript{\rm 1,2}\thanks{ Equal contribution.} ,  
Muhe Cai\textsuperscript{\rm 1,2*}, \hspace{0.1mm}
Jiadong Xu\textsuperscript{\rm 1}, \hspace{0.1mm}
Juan Zhu\textsuperscript{\rm 2},  \hspace{0.1mm}
Hongwei Fan\textsuperscript{\rm 1,2}, \hspace{0.1mm}\\
Yan Shen\textsuperscript{\rm 1,2},  \hspace{0.1mm}
Guanghui Ren\textsuperscript{\rm 3}, \hspace{0.1mm}
Hao Dong~\textsuperscript{\rm 1,2} \hspace{0.1mm} \\
 \textsuperscript{\rm 1}School of Computer Science, Peking University
\textsuperscript{\rm 2} PrimeBot
\textsuperscript{\rm 3} PKU Lab  \\
}

\begin{document}
\maketitle
\begin{strip}
\vspace{-10mm}
    \centering
    \includegraphics[width=1\textwidth]{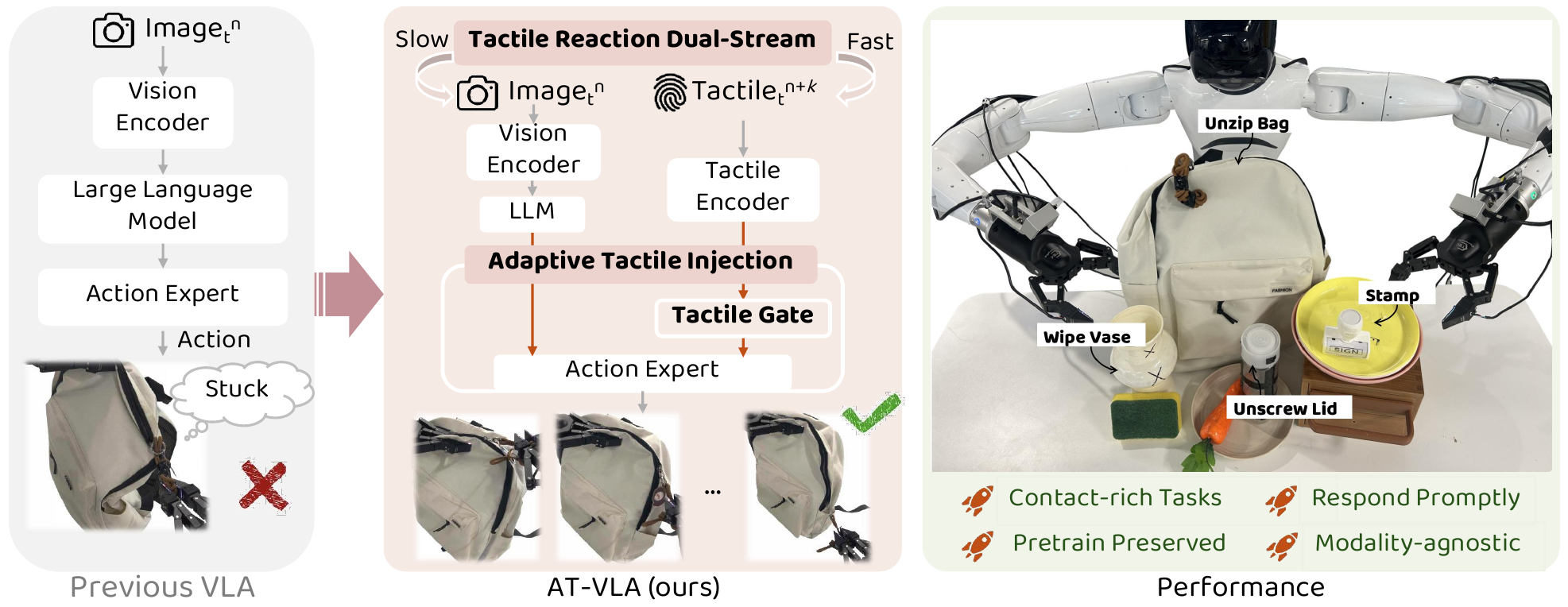}
    \vspace{-7mm}
    \captionof{figure}{
    \textbf{AT-VLA} improves upon previous VLA approaches in contact-rich tasks by introducing Adaptive Tactile Injection, which balances pretrained knowledge with the learning of newly incorporated tactile representations. Furthermore, it enables rapid and accurate action adjustments based on tactile feedback through a Tactile Reaction Dual-Stream Strategy. 
    }  
    \label{fig:teaser}
\end{strip}

\begin{abstract}
Vision-Language-Action (VLA) models have significantly advanced the capabilities of robotic agents in executing diverse tasks; however, they still face challenges in contact-rich manipulation scenarios that require precise physical interactions. 
To address this limitation, recent studies have attempted to incorporate tactile signals during downstream tasks, enabling pretrained VLAs to interpret tactile feedback. 
Nevertheless, introducing new modalities during finetuning, which are rarely present in the pretrain stage, may disrupt the pretrained capabilities of VLAs. 
In addition, the inherently slow inference speed of VLAs hampers real-time responsiveness and limits the effective utilization of tactile feedback for action adjustment.
To overcome these challenges, we propose Adaptive Tactile Vision-Language-Action (AT-VLA), which introduces a novel Adaptive Tactile Injection mechanism. 
This mechanism dynamically determines the appropriate timing and locations for tactile injection, incorporating only when it significantly contributes to action generation, thereby minimizing interference with pretrained representations.
Furthermore, to enable rapid and accurate tactile responses, we propose a Tactile Reaction Dual-Stream mechanism, which decouples sensory processing into a slow visual-language stream for low-frequency perceptual reasoning and a fast tactile control stream for high-frequency physical interaction understanding, achieving real-time close-loop responses within 0.04 s.
Real-world experiments thoroughly validate the effectiveness of AT-VLA in contact-rich manipulation tasks.
The project page is available at: https://sites.google.com/view/at-vla.
\end{abstract}
\section{Introduction}
\label{sec:intro}
The development of Vision-Language-Action (VLA) models~\cite{black2024pi_0,wen2025dexvla,o2024open,brohan2022rt,yue2024deer,wen2025tinyvla,zitkovich2023rt,intelligence2504pi0} has greatly accelerated the progress toward generalist robotic agents.
Empowered by large-scale manipulation datasets~\cite{o2024open,wu2024robomind,bu2025agibot} and the emergence of foundation models~\cite{wang2024qwen2,beyer2024paligemma}, VLAs demonstrate strong abilities that enable robots to ground language in perception and perform diverse tasks.
However, when facing contact-rich manipulation scenarios that require precise understanding of physical interactions, these models remain limited, as they often overlook interaction feedback (\emph{e.g.} tactile signals) that are essential for achieving intricate control and safe interaction with the physical world.


Since pretrained open-source manipulation datasets rarely include tactile information, researchers~\cite{bi2025vla,hao2025tla,yang2025bitla} often address this limitation by incorporating these modalities during downstream tasks finetuning. 
Their primary goal is to enable pretrained models to interpret these new types of sensory input. 
This is typically achieved by using multimodal alignment strategies through representation learning~\cite{cheng2025omnivtla,zhang2025vtla,zhang2025ta,yu2025forcevla}, or leveraging chain-of-thought (CoT)~\cite{huang2025tactile} reasoning to understand these signals.
However, tactile feedback provides fundamentally different types of information compared to the visual or linguistic data used in the pretrained models. 
While these approaches enhance the model's ability to interpret tactile feedback, the direct introduction of these new modalities may disrupt the existing pretrained knowledge, such as visual perception or object grounding.
Furthermore, the relatively slow inference speed of VLA models hampers their ability to react quickly to high-frequency tactile feedback. 
This slow response time reduces the effectiveness of tactile feedback in closed-loop manipulation, where swift adjustments are crucial.
To this end, as illustrated in Fig.~\ref{fig:teaser}, we propose Adaptive Tactile Vision-Language-Action (AT-VLA), which, for the first time, achieves a balance between preserving pretrained capabilities and integrating tactile inputs, while ensuring rapid and effective tactile feedback reactions.

Drawing inspiration from the complementary characteristics of visual and tactile modalities, where vision facilitates contextual localization and tactile provides precise contact feedback, we argue that the model should preserve its pretrained VLA structure in non-contact phases while introducing tactile feedback only upon contact to better utilize pretrained representations.
To realize this, we introduce an \textbf{Adaptive Tactile Injection} mechanism that determines when and where to incorporate tactile signals.
Specifically, a learnable \textit{Tactile Gate} is designed to automatically modulate the contribution of each modality across different manipulation phases, determining whether tactile features should be fed into the action expert for action generation.
To accommodate different gate states, we further propose an \textit{Adaptive Cross Attention} mechanism within the action expert module, which dynamically adjusts its cross-attention behavior by conditionally switching the query source.
This mechanism integrates tactile information when the gate is activated, without the need to modify the model structure or feature dimensionality compared to the inactive state.
By doing so, we minimize the impact of tactile signals on the vanilla model, thus retaining its robust action generation capability, such as target object localization.

Furthermore, since high-frequency tactile feedback requires rapid action adjustment, we design a \textbf{Tactile Reaction Dual-Stream} mechanism to ensure the model’s real-time responsiveness when tactile gate is activated. 
Specifically, we decouple sensory processing into two streams of different frequency: a \emph{slow stream}, where visual and language inputs are processed through a large Vision-Language Model (VLM) to leverage the pretrained visual perception and localization ability at a low frequency; and a \emph{fast stream}, where tactile feedback is continuously fed into the action expert at a high frequency. 
The action expert integrates heterogeneous modality features operating at different frequencies, conditioning each action prediction on both the latest tactile feedback and the most recent vision-language reasoning output that falls within the same action chunk horizon.
This design enables a closed-loop control within 0.04\,s inference speed, ensuring timely responses to tactile feedback. 

Our main contributions are as follows: 
1) We propose Adaptive Tactile Injection, making the first attempt to balance pretrained knowledge with the learning of newly introduced tactile representations.
2) Leveraging the high-frequency nature of tactile feedback that demands rapid response, we design a Tactile Reaction Dual-Stream mechanism to ensure accurate and timely action adjustments during contact.
3) AT-VLA achieves superior manipulation performance on real-world contact-rich tasks compared with other SOTA VLA or tactile-based policies.
4) Unlike prior tactile-based policies that heavily rely on tactile inputs, AT-VLA, although trained with tactile feedback, maintains strong performance even in the absence of tactile signals during inference.
Benefiting from the adaptive training strategy, AT-VLA generalizes effectively across diverse sensory conditions while maintaining performance comparable to the vanilla VLA model, demonstrating strong modality-agnostic robustness in real-world circumstances where tactile signals may be unstable or unavailable.


\section{Related Work}
\begin{figure*}[t]
\centering
\includegraphics[width=0.995\textwidth]{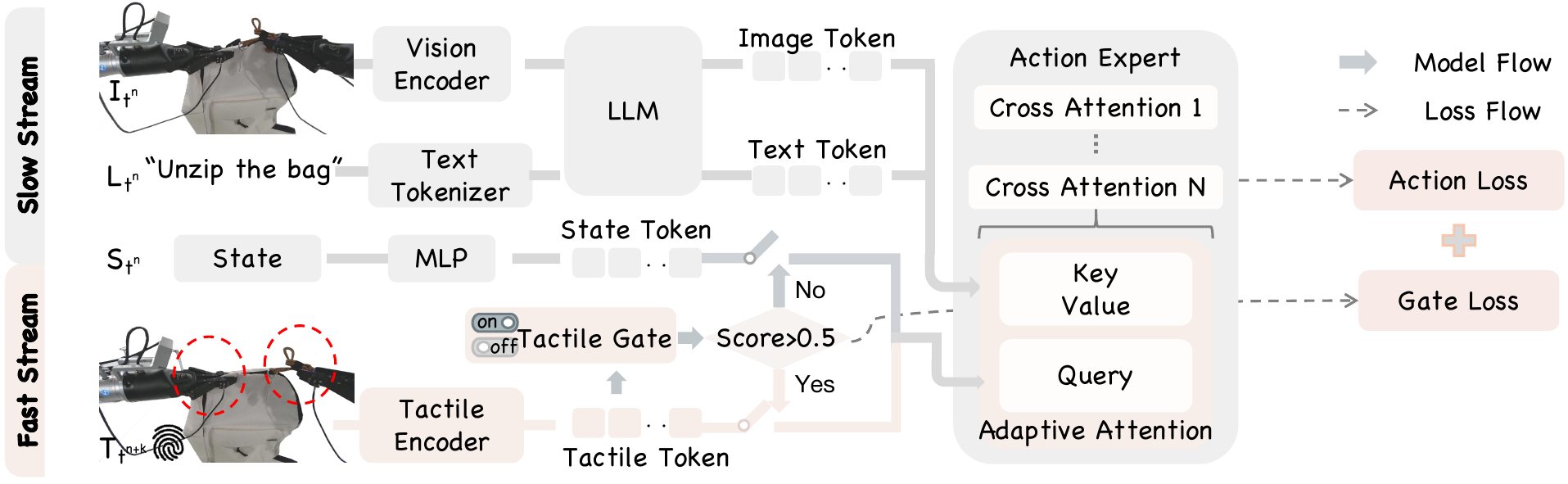} 
\caption{
\textbf{Framework of AT-VLA.} The tactile gate adaptively determines whether tactile tokens should be used as conditional inputs for action generation within the Action Expert module. When the tactile gate is inactive, all input modalities of the Action Expert operate at the same frequency. When activated, the tactile signal is processed at a higher frequency to enable rapid and precise action adjustments.
}
\label{fig:method}
\vspace{-0.4cm}
\end{figure*}
\subsection{Vision Language Action Model}
Rapid progress in Vision Language Models (VLMs)~\cite{alayrac2022flamingo,liu2023visual,achiam2023gpt} has laid the foundation for the emergence of Vision Language Action (VLA) Models. 
By integrating environment perception, semantic reasoning, and action generation within a unified framework, VLAs~\cite{kim2024openvla,li2024cogact} demonstrate strong potential for task generalization and reliable manipulation.
One common approach for action generation is to adopt diffusion-based models to formulate a conditional denoising process.
For example, $\pi_{0}$~\cite{black2024pi_0} leverages flow matching (a specific instantiation of diffusion) to enhance temporal resolution and highlights the importance of effective training strategies by introducing a componentized architecture, where outputs from a visual-language model serve as preconditions for the diffusion-based action transformer.
In this paper, following the vanilla VLA GO-1~\cite{bu2025agibot}, we adopt this  manner to formulate the action modeling.
Moreover, to enhance inference speed, several VLA~\cite{cui2025openhelix,chen2025fast,lin2025onetwovla,chen2025ac,li2025object,liu2025hybridvla,li2024manipllm,liu2024robomamba} approaches adopt a dual-system strategy for action generation, where the slow system targets on high-level Vision-Language Model reasoning and the fast system acts as low-level policy.
For instance, Gr00t-N1~\cite{bjorck2025gr00t} employs a fast visual stream alongside a slower semantic reasoning stream to preserve high-level planning ability and accelerate low-level action prediction.
Different from these approaches, which use visual or point cloud data as the fast stream, we adopt high-frequency tactile feedback as the fast stream, enabling rapid reaction to contact events and safer physical interactions.

\subsection{Interaction Feedback Policies}
Vision is fundamental to environmental perception, while other interactive modalities~\cite{funk2025importance,funk2024evetac,bohm2024matters,zhu2025touch,huang2025vt,liu2025mla}, such as joint torque, force, and end-effector tactile signals, are equally crucial for precise, closed-loop manipulation.
Specifically, RDP~\cite{xue2025reactive} utilizes PCA-reduced marker offset as tactile input for latent space decoder, achieving high-frequency feedback adjustment.
TLA~\cite{hao2025tla} relies on tactile and language input and works well in insertion tasks trained on simulation data.
TA-VLA~\cite{zhang2025ta} identifies that incorporating a tactile adapter within the decoder yields better performance than placing it in the encoder.
In contrast to those that focus on enabling the model to interpret the semantic meaning of tactile feedback but may compromise the model's original visual perception ability, we carefully manage the balance between pretrained knowledge and newly introduced tactile feedback, ensuring reliable manipulation in both non-contact and contact-rich stages.
Moreover,  compared to current tactile gate works~\cite{hansen2022visuotactile,li2025adaptive,he2025foar}, our method is designed to accommodate the characteristics of pretrained VLAs. 
When tactile gate is activated, these works typically incorporate tactile features by augmenting token sequence. 
In contrast, we observe that naively appending previously unseen modality may disrupt VLA’s pretrained token sequence modeling. 
In addition, these works typically process all modalities at the same operation frequency. However, due to the inference latency of VLAs, we decouple different modality processing frequencies.
\section{Method}
\label{sec:formatting}

\subsection{Framework of AT-VLA}
\label{sec:framework}
\noindent\textbf{Problem Formulation.}
As shown in Fig.~\ref{fig:method}, the policy $\pi_{\theta}$ takes as input the image observations $I = \{I_{h}, I_{r}, I_{l}\}$ from the head camera, right wrist camera, and left wrist camera, respectively; the language instruction $L$; the tactile feedback $T$; and the robot's proprioceptive state $S$. 
For the tactile feedback, we extract the resultant force from the tactile sensors, which contains both the 3D normal and 3D tangential components of contact forces. 
The policy then generates an action chunk $A$, representing the 14-DoF end-effector pose for both arms: 
\[
A = \pi_{\theta}(I, L, T, S).
\]

\noindent\textbf{Model Instantiation.}
AT-VLA is designed to be general and modular, allowing most VLAs to be integrated with minimal modification. 
For a specific instantiation, we employ the pretrained GO-1~\cite{bu2025agibot} as the vanilla VLA, since it is trained on the same robot hardware (Genie~1 from AgiBot) as ours and demonstrates strong performance across diverse tasks. 
GO-1 is pretrained in the AgiBot world dataset~\cite{bu2025agibot} and adopts Intern-VL-2B~\cite{chen2024internvl} as its base vision-language model (VLM). It utilizes a DiT~\cite{peebles2023scalable} module as the action expert. 
We inherit both its model architecture and its action generation pipeline, where the actions are supervised by the action loss $\mathcal{L}_{a}$. 
To enable the model to handle contact-rich tasks, we introduce an additional tactile encoder. 
The tactile encoder is a lightweight module composed of several MLP layers, designed to ensure fast inference while efficiently processing tactile signals.

\subsection{Adaptive Tactile Injection}
\label{sec:AFI}
\textbf{Intuition.}
\begin{figure}[t]
\centering
\includegraphics[width=0.49\textwidth]{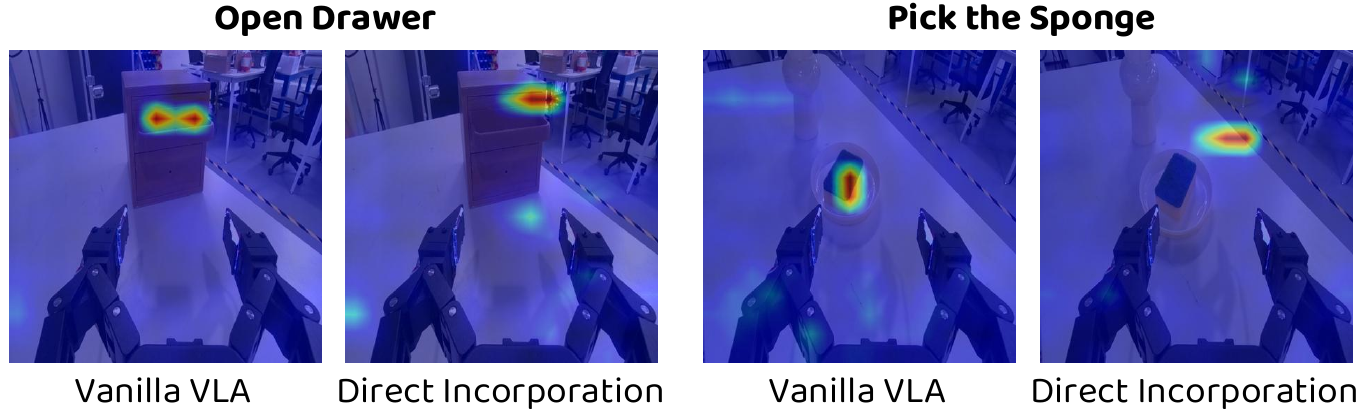} 
\vspace{-0.3cm}
\caption{
\textbf{Intuition.} We visualize the attention maps in the Action Expert module to examine how the model’s attention distribution and action reasoning vary across downstream finetuning strategies, contrasting settings with and without tactile feedback. }
\label{fig:mot}

\vspace{-0.1cm}
\end{figure}
We conduct a simple experiment that compares two training strategies for learning downstream manipulation tasks: (1) a typical fine-tuning strategy, which fine-tunes only the action expert of the vanilla VLA model without tactile input; and
(2) a direct way to incorporate tactile feedback, which introduces tactile signals processed by the tactile encoder and fine-tunes only the action expert to condition on both the tactile tokens and the vision-language tokens.
Surprisingly, the latter not only fails to improve performance but even causes noticeable inaccuracies in grasp localization.
A detailed analysis of the attention maps within the action expert (Fig.~\ref{fig:mot}) reveals that the additional tactile input shifts the model’s attention distribution away from the target object towards surrounding regions. 
We hypothesize that the addition of token sequences from newly introduced modalities can interfere with the pretrained model’s perceptual focus.
 Further quantitative results supporting this observation are provided in Tab.~\ref{tab:ablation}.
This observation motivates us to consider that balancing pretrained knowledge with the learning of newly introduced tactile feedback must be carefully managed.


Therefore, to address these issues, we propose the \emph{Adaptive Tactile Injection} module, which dynamically controls when and where tactile feedback is injected and enables the action expert to flexibly handle different gate states through adaptive cross-attention.
Given the complementary nature of vision and tactile sensing, where vision provides contextual information for localization and tactile sensing offers detailed feedback on physical contact, the key insight is to enable the model to retain the behavior of the vanilla VLA in non-contact phases while leveraging tactile information only during contact.

\noindent\textbf{Tactile Gating.}
To realize it, we first design a learnable Tactile Gate mechanism that automatically determines when tactile feedback should be incorporated, which is when the robot is in contact with an object.
Specifically, after extracting the tactile token $\mathbf{z}_{T}$ using the tactile encoder, we introduce a lightweight Tactile Gating Network composed of MLP layers to classify whether the current tactile signal indicates contact or non-contact.
The network takes the tactile token as input and outputs a score representing the contact state.
For supervision, we manually annotate the training episodes by assigning a label of 0 to non-contact frames and 1 to contact frames, and adopt binary cross-entropy gating loss $\mathcal{L}_{g}$ to supervise.
When the score exceeds the threshold (e.g., 0.5), the tactile gate is activated.
This gate not only helps in discriminating different contact stages, but also forces the model to better comprehend tactile signals.

\noindent\textbf{Adaptive Cross Attention.}
With the tactile gate to determine when to incorporate tactile feedback, the action expert’s architecture must be able to handle inputs under both states of the tactile gate, whether or not tactile features are present. 
To realize this, we propose \emph{Adaptive Attention} in cross-attention module, which dynamically processes different input modalities of the action expert as conditions for action generation.
Specifically, in the cross attention module of the action expert in the vanilla VLA, the image token $\mathbf{z}_{I}$ and text token $\mathbf{z}_{L}$ serve as the key and value, while the state token $\mathbf{z}_{S}$ is used as the query.  
To maintain consistency with the vanilla VLA and preserve its pretrained representations, the state token is used as the query when the tactile gate is inactive, and is replaced by the tactile token $\mathbf{z}_{T}$ when the gate is active.
In this way, when there is no contact, the model’s input and architecture remain identical to those of the vanilla VLA. Once contact occurs, however, the model starts interpreting the tactile feedback as action generation conditioning. 
The model thus preserves pretrained representations to the greatest extent, maintaining action behaviors derived from strong visual perception and localization capabilities, such as approaching target objects.



\subsection{Effective Tactile Reaction Dual-Stream}
\label{sec:dual}
In Sec.~\ref{sec:AFI}, our model adaptively determines \textit{when} and \textit{where} to inject tactile feedback, balancing the preservation of pretrained knowledge and learning of newly introduced modality. 
In this section, motivated by the high-frequency nature of tactile inputs requiring rapid action adaptation, we explore \textit{how} to enable the model to respond effectively when the tactile gate is activated.
We decompose the reaction capability into two aspects: (1) rapid reaction. The model should quickly adjust its predicted actions based on incoming tactile feedback to ensure safety and precision, and (2) tactile understanding. The model should learn to interpret the meaning of tactile signals and modify its actions accordingly.

\textbf{For a rapid reaction}, we define the input modalities process into two streams: a slow stream operates at a lower inference speed to interpret visual and language input via large vision language model; a fast stream operates at a high inference speed to interpret tactile feedback. 
Slow stream is responsible for task understanding and visual perception, producing a comprehensive output in the form of latent features serving as the key and value in the cross attention of action expert module, whereas fast stream processes tactile feedback that requires closed-loop real-time response serving as the cross attention query conditions.
Therefore, the input of action expert module is at asynchronous frequency and with heterogeneous modality. 

Building on previous action chunking strategies, the visual and language observation at time step $t_{n}$ can provide guidance for a future horizon of action steps ($t_{n}$:$t_{n+H}$).
Consequently, slow stream's output serves as a latent condition that temporally guides action generation across the following $H$ time steps. 
In contrast, fast stream focuses on generating executable actions in real time. 
At each time step, action expert leverages the most recent tactile feedback $t_{n+k, 0<k<H}$ to generate actions, while conditioning on the periodically updated output from the slow stream, i.e., the slow stream’s output at time $t_n$.
During training, the frequency ratio between the fast and slow streams is randomly set to $h\!:\!1, \text{ where } 1 < h < H$, meaning that the fast stream’s input can be ahead of the slow stream by at most the action chunk horizon.
Note that, when the tactile gate is inactive, the input of the action expert module is at synchronous frequency.
In this way, the fast stream leverages the pretrained representations from the slow stream to ensure reliable visual perception, while simultaneously reacting to tactile dynamics within a 0.04s inference speed.

Furthermore, to advance the fast stream with an accurate reaction to tactile input, we aim to enhance the model towards \textbf{a deeper understanding of tactile feedback} along with robotic physical interactions. 
In the previously introduced Tactile Gating Mechanism, the model learns to interpret whether contact is established based on the tactile token, improving its awareness of contact states and transitions during manipulation.
This design encourages the model to develop a more comprehensive representation of physical dynamics and tactile semantics, bridging instantaneous contact perception and predictive interaction reasoning.

\subsection{Training Objectives and Inference Pipeline}

All objectives are trained simultaneously, under the overall supervision 
\[
\mathcal{L} = \mathcal{L}_{a} + \lambda_1*\mathcal{L}_{g} 
\]
, $\lambda_1$ is set to 0.01 to balance different losses' scale.
By doing so, AT-VLA is able to preserve pretrained knowledge and response to tactile feedback promptly and accurately.

During inference, when the tactile gate is inactive, the model is identical to the original VLA, the input of fast and slow stream is at the same frequency, the attention query in the action expert module is same as the vanilla VLA.
When the tactile gate is active, asynchronous frequency of slow and fast starts. 
Inspired by previous work~\cite{bjorck2025gr00t,chen2025fast}, we set the frequency ratio between fast stream and slow stream to 3:1 to ensure a balance of efficiency and performance, which means that the fast stream infers three times consecutively after a single inference of the slow stream.
At the same time, the attention queries in the action expert module are switched to tactile tokens, which serve as the conditioning for action generation.

\section{Experiment}

\begin{table*}[t]
\centering
\renewcommand{\arraystretch}{1.2}
\setlength{\tabcolsep}{4pt}
{\normalsize
\caption{\textbf{Evaluation in contact-rich tasks.} We report the success rate of each subtask, reflecting the progress. }
\label{tab:main}
\resizebox{1.0\textwidth}{!}{
\begin{tabular}{l|ccccc|cccc}
\toprule
& \multicolumn{5}{c|}{\textbf{Unzip Bag}} & \multicolumn{4}{c}{\textbf{Stamp}}\\
\cmidrule(lr){2-6} \cmidrule(lr){7-10}
Method & Left Grasp $\rightarrow$ & Right Grasp $\rightarrow$ & Unzip Half $\rightarrow$ & Unzip Full & \textbf{Overall} & Right Grasp $\rightarrow$& Stamp $\rightarrow$& Place & \textbf{Overall} \\
\midrule
GO-1 &  \textbf{1.0}&\textbf{0.87} & \textbf{0.87} & 0.20 & 0.20&\textbf{1.0}&0.80&0.13&0.13\\
$\pi_{0.5}$ & \textbf{1.0}&0.67&0.67&0.0&0.0&\textbf{1.0}&0.53&0.20&0.20\\
\textbf{AT-VLA (Ours)} & \textbf{1.0}&0.80&0.80&\textbf{0.33}&\textbf{0.33}&0.90&0.87&\textbf{0.46}&\textbf{0.46}\\\hline
VTLA & - & - & 0.20& 0.00   &-&-&\textbf{0.93} & 0.13&-\\
RDP & - & - & \textbf{0.87} & 0.06 &-&-&0.40&0.40&-\\
\hline
\hline
Method & Left Grasp $\rightarrow$ & Right Grasp $\rightarrow$ & Wipe Half $\rightarrow$ & Wipe Full & \textbf{Overall} & Left Grasp $\rightarrow$ & Right Grasp $\rightarrow$ & Rotate & \textbf{Overall} \\
\midrule
GO-1 & \textbf{1.0}&\textbf{0.93}&0.30&0.07&0.07&\textbf{1.0}&\textbf{0.87}&0.27&0.27 \\
$\pi_{0.5}$ &\textbf{1.0 }&0.73&0.40&0.33&0.33&0.93&0.80&0.46&0.46\\
\textbf{AT-VLA (Ours)} & \textbf{1.0}&0.80&0.80&\textbf{0.67}&\textbf{0.67}&0.90&0.87&0.53&\textbf{0.53}\\\hline
VTLA & -&-& \textbf{0.93}&0.60&-&-&-&0.80&-\\
RDP & -&-& 0.80&0.33&-&-&-&\textbf{0.87}&-\\
\bottomrule
\end{tabular}
}}
\end{table*}

\subsection{Setup}
\noindent\textbf{Hardware.}
We use the hardware, AgiBot Genie1, featuring dual 7-DoF arms, equipped with a front-view camera and two cameras mounted on the wrist.
For tactile feedback, a gripper equipped with tactile sensors from Xense Robotics is utilized.
We use the VR headset control for teleoperation.

\noindent\textbf{Task Definition.}
We evaluate our model on four contact-rich tasks and two non contact-rich tasks. For each task, we collect 30-50 demonstrations and test in 15 trials.
\textbf{Contact-rich tasks}: 
a). \textit{Unzip Bag}. The robot is required to open a bag by unzipping its zipper. During this process, it must continuously adjust its trajectory to follow the curved path of the zipper. Failure to do so may cause the zipper to get stuck or jammed. 
b). \textit{Stamp}. The robot is required to stamp within a designated region. It must recognize when the stamping action is complete; otherwise, excessive downward actions could cause the end effector to collide with the desk surface.
c). \textit{Wipe Vase}. The robot needs to wipe the surface of a vase, adapting its motion to the vase’s curved geometry. Insufficient compliance could result in collisions with the neck of the vase.
d). \textit{Unscrew Lid}. The robot is required to rotate a lid to open a container. This task demands precise force and motion coordination to ensure smooth rotation without slipping.
\textbf{Non contact-rich tasks}: 
a). \textit{Pick and Place}. The robot needs to pick a carrot and place it in the plate.
b). \textit{Open Drawer}. The robot needs to grasp the handle of a drawer and open it.

\subsection{Contact-rich Task Evaluation}
\label{sec:mainexp}

\noindent\textbf{Baselines.}
We compare with four SOTA baselines:
\begin{enumerate}
   
\item{\textbf{GO-1}~\cite{bu2025agibot}}
 serves as the vanilla model of our approach and is pretrained on the same dataset as ours. It can reflect how much improvement our method achieves.  
 
\item{\textbf{$\mathbf{\pi_{0.5}}$}~\cite{intelligence2504pi0}} 
is a state-of-the-art VLA model consisting of both pretraining and post-training stages.  

\item{\textbf{VTLA}~\cite{zhang2025vtla}}
is a vision-language-action (VLA) model that incorporates tactile input. It employs Qwen2-VL~\cite{wang2024qwen2} as the VLM backbone and adds an additional ViT~\cite{dosovitskiy2020image} to extract tactile features, where tactile information is represented as visual-tactile images. Since the official implementation is not publicly available, we replicate it based on the DexVLA~\cite{wen2025dexvla} codebase, which also builds upon Qwen2-VL.  

\item{\textbf{RDP}~\cite{xue2025reactive}} 
is built upon Diffusion Policy~\cite{chi2025diffusion} and takes 2D tactile marker inputs. It applies PCA for dimensionality reduction before extracting tactile features.

\end{enumerate}
Note that,
1) GO-1 and $\pi_{0.5}$, which have pretrained weights, are trained on the same downstream dataset as ours but without tactile input. 
2) In contrast, VTLA and RDP, which do not have pretrained models on large-scale datasets, are trained only on the subset of our downstream tasks corresponding to the contact-rich manipulation phases.
We found that training them on the full sequence often leads to failures during the grasping stage, which makes it difficult to reveal their core capability—namely, reacting to tactile feedback. Therefore, we remove the perception and grounding requirements for these two models to better isolate their ability to respond to tactile feedback.
In practice, during testing, we manually place the robot in an ideal initial configuration (\emph{e.g.,} already grasping the stamp) to evaluate these two models’ capability in completing the contact-rich portion of the task (\emph{e.g.,} stamping).

\begin{figure*}[t]
\centering
\includegraphics[width=1.0\textwidth]{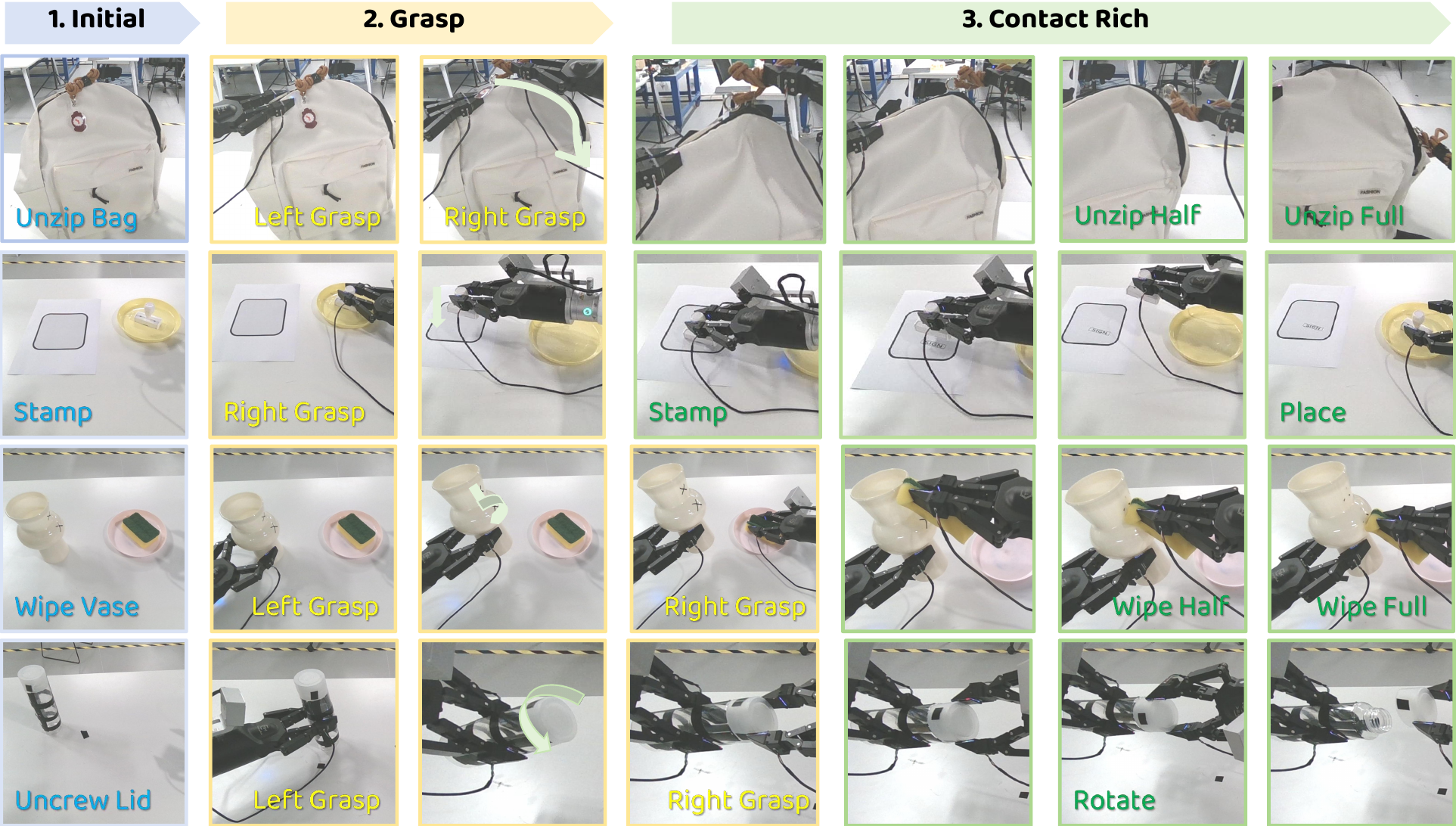} 
\caption{
\textbf{Visualization.} We visualize the execution progress of four typical contact-rich tasks.
}
\label{fig:viz1}
\end{figure*}

\noindent\textbf{Result Analysis.}
As shown in Table~\ref{tab:main}, our model outperforms all baseline methods. 
The execution progress is shown in Fig.~\ref{fig:viz1}.
Compared with state-of-the-art VLA models GO-1 and $\pi_{0.5}$, which are trained without tactile feedback, our model demonstrates comparable performance during the pre-contact manipulation phase, indicating that it effectively preserves the pretrained knowledge in aspects such as visual grounding and semantic reasoning required for reliable object grasping.
During the contact-rich stage, AT-VLA achieves an improvement over them, clearly demonstrating the necessity of tactile signals for complex manipulation tasks. 
In contrast, these models usually get stuck in stages, such as getting stuck by the zipper, or stamping on the desk too much, or getting stuck by the protuding surface.

Furthermore, when compared with policies that incorporate tactile feedback like VTLA and RDP, our model still achieves superior performance in contact-rich phase manipulation, validating the effectiveness of our model to promptly and accurately adjust actions based on tactile feedback.
The only exception is Unscrew Lid task, where our model performs slightly worse than them.
This is mainly because, for baseline policies, we manually set the robot in an ideal grasping pose before rotating the lid, ensuring a stable grasp during the process. 
In contrast, our model, although capable of grasping the lid, does not always guarantee a sufficiently firm grip, occasionally leading to failure cases where the gripper slips during unscrewing.

\subsection{Modality-agnostic Evaluation}
\label{sec:modexp}
\begin{table}[h]
\centering
\renewcommand{\arraystretch}{1.0}
\setlength{\tabcolsep}{8pt}
\caption{\textbf{Modality-agnostic evaluation.}The AT-VLA variants with (w/.) and without (w/o.) tactile input share identical model weights, differing only in whether tactile information is provided during inference. The former serves as upper bound.}
\label{tab:mod}
\resizebox{0.48\textwidth}{!}{
\begin{tabular}{l|cc|c|c}
\toprule
Method & \textbf{Pick Place} &  \textbf{Open Drawer}&  \textbf{Stamp} & AVG.\\
\midrule
GO-1 &\textbf{1.0} & \textbf{0.93}&0.13&0.68\\ 
$\pi_{0.5}$&\textbf{1.0}&\textbf{0.93}& \textbf{0.20} & \textbf{0.70}\\
\textbf{AT-VLA w/o.} & \textbf{1.0}& \textbf{0.93}&\textbf{0.20}& \textbf{0.70}\\\midrule
\textcolor{gray}{AT-VLA w/.} & \textcolor{gray}{1.0}& \textcolor{gray}{0.93}& \textcolor{gray}{0.46} & \textcolor{gray}{0.79}\\
\bottomrule
\end{tabular}
}
\end{table}
In this section, we aim to evaluate the modality-agnostic robustness of our model.
That is, how well it performs when trained on downstream tasks with tactile feedback but deployed without tactile signals during inference, which is crucial for real-world robotic applications where sensor failures or missing modalities are common.

As shown in Table~\ref{tab:mod}, we compare our method with two VLA baselines, GO-1 and $\pi_{0.5}$, which are also without tactile feedback input.
Our two model alternatives, AT-VLA w/. and AT-VLA w/o., share the same model weights trained with tactile input; however, the former performs inference with tactile feedback which serves as an upper bound, while the latter infers without it.
When compared with the VLA baselines, AT-VLA (w/o tactile) achieves comparable performance on two non contact-rich tasks, Pick-and-Place and Open Drawer.
Thanks to the Adaptive Tactile Injection mechanism, our model can still generate reliable actions even when tactile feedback is absent at inference, demonstrating strong modality-agnostic and robustness to missing sensory modalities.

For contact-rich manipulation, we select Stamp task as a representative example, since it involves intensive physical interaction while remaining executable by base VLA baselines with relatively high safety.
Interestingly, AT-VLA (w/o tactile) even shows a slight performance improvement over its vanilla model GO-1.
This can be attributed to the fact that, although tactile feedback is absent during inference, the model is trained with tactile input. 
Consequently, it has implicitly learned richer contact dynamics and cross-modal correlations during training, allowing it to infer approximate tactile cues from visual features at test time. 
\subsection{Ablation Study}
\label{sec:ablaexp}
\begin{table*}[h]
\centering
\renewcommand{\arraystretch}{1.0}
\setlength{\tabcolsep}{2pt}
\small
\caption{\textbf{Ablation study.} Each variant selectively removes or changes components to assess their contributions.}
\label{tab:ablation}
\resizebox{1.0\linewidth}{!}{
\begin{tabular}{l|cccc|ccc|cccc|c}
\toprule
& \multicolumn{4}{c|}{\textbf{Components}} & \multicolumn{3}{c|}{\textbf{Tactile Format}}& \multicolumn{4}{c|}{\textbf{Tasks}}\\
\cmidrule(lr){2-5} \cmidrule(lr){6-8} \cmidrule(lr){8-12}
 & \makecell{Tactile\\Gate} & \makecell{Adaptive\\Cross Attention}& \makecell{Direct\\Incorporation} & \makecell{Reaction\\Dual-Stream}  & \makecell{V-T\\Image} & \makecell{Marker\\2D} & \makecell{Force\\6D} & \makecell{Unzip\\Bag}& \makecell{Stamp\\ }& \makecell{Wipe\\Vase }& \makecell{Unscrew\\Lid } & AVG. \\
\midrule
Ex0 & - & - &- & - & - & - & -& 0.20&0.33&0.07&0.27 &0.22\\
Ex1 & - & -& \checkmark& - & - & - & \checkmark& 0.07&0.13&0.07&0.20 &0.13\\
Ex2 & \checkmark & \checkmark& - & - &- & - & \checkmark& 0.27&0.40&0.53&0.33&0.39 \\\midrule
Ex3 (Ours) & \checkmark &\checkmark& -  &\checkmark  &- & - & \checkmark& \textbf{0.33}&\textbf{0.46}&\textbf{0.67}&\textbf{0.53}&\textbf{0.50}\\\midrule
Ex4 &-& - &\checkmark &- &- & \checkmark &-& 0.00&0.13&0.07&0.00&0.05\\
Ex5 &\checkmark &\checkmark& -  &\checkmark &- & \checkmark &-& 0.27 & 0.33&0.27&0.40& 0.32\\
\midrule
Ex6 & -& - & \checkmark  & -&\checkmark&-&-& 0.00&0.00&0.07&0.00&0.02 \\
Ex7 & \checkmark & \checkmark& -  & \checkmark&\checkmark&-&-& 0.27&\textbf{0.46}&0.47&0.40&0.40 \\

\bottomrule
\end{tabular}
}
\end{table*}
\subsubsection{Contribution of Each Component}
From Rows Ex0–Ex3 in Tab.~\ref{tab:ablation}, we verify the contribution of each proposed component across four contact-rich tasks: 

\noindent\textbf{Ex0: Vanilla VLA.}  
This setting corresponds to the vanilla VLA model (\emph{i.e.,} GO-1~\cite{bu2025agibot}), which achieves an average success rate of 0.22. 

\noindent\textbf{Ex1: Direct Tactile Incorporation.}  
Compared to Ex0, we introduce tactile input but in a direct manner, consistent with the experiments discussed in the intuition analysis in Sec.~\ref{sec:AFI}.
Specifically, since there is no gating mechanism, the queries in the Adaptive Cross Attention module are conditioned on the tactile features extracted from the tactile encoder throughout the entire manipulation process.
However, this configuration even results in a lower success rate than the vanilla VLA, with a decrease of 9\%.  
Most failures occur during object grasping, indicating degraded visual grounding and perception ability.  
This phenomenon is consistent with our intuition analysis in Sec.~\ref{sec:AFI}. 

\noindent\textbf{Ex2: With Tactile Gate.}  
Compared to Ex1, we further incorporate a Tactile Gate.
Together with Adaptive Cross Attention, it enables the attention query to switch to tactile tokens only when the gate is activated; otherwise, the attention query remains identical to that of the vanilla VLA. 
Note that, in this setting, the visual and tactile inputs are processed at the same frequency.  
This configuration achieves a significant 17\% improvement over the vanilla VLA, demonstrating the effectiveness of the tactile gate in preserving the pretrained knowledge of the model.

\noindent\textbf{Ex3: With Reaction Dual-Stream.}  
Compared to Ex2, we adopt the Tactile Reaction Dual-stream, setting different input frequencies for visual and tactile modalities.  
This setting yields an improvement of 11\%, highlighting the necessity of rapid tactile feedback responses. In \textit{Unzip} task, for instance, where delayed reactions may cause the zipper to jam and hinder successful completion.

\subsubsection{Different Tactile Format}
We compare the effectiveness of incorporating tactile information in different formats, including visual-tactile images (V-T image in Tab.~\ref{tab:ablation}), marker 2D, and force 6D.  
Specifically, the \textit{visual-tactile image} contains contact geometry by mapping tactile deformation into an RGB image of size $H \times W \times 3$, $H$ and $W$ are the size of tactile sensor. We adopt a pretrained tactile encoder Sparsh~\cite{higuera2024sparsh} to extract its feature, serving as tactile token in Fig.~\ref{fig:method};  
the \textit{marker 2D} format represents the displacement of surface markers as a $N \times 2$ array, where $N$ denotes the number of tracked markers. We initialize a lightweight encoder consisting of several MLP and non-linear activation layers to extract its feature;  
and the \textit{force 6D} format represents the resultant contact force, composed of a 3D tangential force vector and a 3D normal force vector, jointly capturing the surface’s shear and normal interactions during contact.

Comparing Ex1, Ex4, and Ex6 with the vanilla VLA (Ex0), these variants all introduce tactile modalities—namely, 6D force, 2D marker, or visual-tactile images. The extracted tactile tokens are fed into the cross-attention queries of the action expert module throughout all manipulation stages, which leads to a significant performance degradation. 
This observation further confirms that directly incorporating newly added modality tokens can disrupt the pretrained knowledge, resulting in failures even in basic manipulation skills such as object grasping.

In contrast, adopting our proposed method to incorporate tactile input leads to substantial performance improvements—Ex5 outperforms Ex4 by 27\%, and Ex7 surpasses Ex6 by 38\%—demonstrating the robustness of our framework across different tactile formats.
Moreover, we observe that the \textit{force 6D} modality achieves the best performance, followed by \textit{marker 2D}, and then the \textit{visual-tactile image}.
We hypothesize that higher-dimensional tactile inputs may excessively perturb the pretrained representation space, as they introduce a larger number of tactile tokens. 
This highlights the importance of maintaining an appropriate balance between the influence of the newly added tactile modality and that of the pretrained vanilla models.

\section{Conclusion}
In summary, AT-VLA introduces an adaptive framework that seamlessly integrates tactile sensing into vision-language-action models. Through the Adaptive Tactile Injection mechanism, AT-VLA dynamically balances pretrained visual-language knowledge with newly learned tactile representations, preserving model integrity while enhancing action precision. 
The Tactile Reaction Dual-Stream mechanism further enables rapid, high-frequency tactile responses by decoupling slow perceptual reasoning from fast physical control, achieving closed-loop reaction within 0.04 s. 
Extensive real-world experiments demonstrate that AT-VLA significantly outperforms state-of-the-art VLA and tactile-based policies in contact-rich manipulation tasks.
These results highlight AT-VLA’s strong adaptability and modality-agnostic robustness, making it a promising step toward more responsive and generalizable multimodal robotic systems.

\section*{Acknowledgments}
This project was supported by National Youth Talent Support Program (8200800081) and National Natural Science Foundation of China (62376006).
{
    \small
    \bibliographystyle{ieeenat_fullname}
    \bibliography{main}
}


\end{document}